\documentclass[journal]{IEEEtran}

\usepackage[pdftex]{graphicx}

\ifCLASSINFOpdf
\else
\fi
\usepackage{amsmath}

\usepackage{multirow}

\begin{document}
%
\title{Learning Multilayer Channel Features for Pedestrian Detection}
%
%
%


\author{Jiale~Cao,
        Yanwei~Pang,
        and~Xuelong~Li

\thanks{Y. Pang and J. Cao are with the School of Electronic Information Engineering, Tianjin University, Tianjin 300072, China. e-mails: \{pyw,connor\}@tju.edu.cn}

\thanks{X. Li is with the Center for OPTical IMagery Analysis and Learning (OPTIMAL), State Key Laboratory of Transient Optics and Photonics, Xi'an Institute of Optics and Precision Mechanics, Chinese Academy of Sciences, Xi'an 710119, Shaanxi, P. R. China. e-mail: xuelong\_li@opt.ac.cn.}}

\maketitle

\begin{abstract}
Pedestrian detection based on the combination of Convolutional Neural Network (i.e., CNN) and traditional handcrafted features (i.e., HOG+LUV) has achieved great success. Generally, HOG+LUV are used to generate the candidate proposals and then CNN classifies these proposals. Despite its success, there is still room for improvement. For example, CNN classifies these proposals by the full-connected layer features while proposal scores and the features in the inner-layers of CNN are ignored. In this paper, we propose a unifying framework called Multi-layer Channel Features (MCF) to overcome the drawback. It firstly integrates HOG+LUV with each layer of CNN into a multi-layer image channels. Based on the multi-layer image channels, a multi-stage cascade AdaBoost is then learned. The weak classifiers in each stage of the multi-stage cascade is learned from the  image channels of corresponding layer. With more abundant features, MCF achieves the state-of-the-art on Caltech pedestrian dataset (i.e., 10.40\% miss rate). Using new and accurate annotations, MCF achieves 7.98\% miss rate. As many non-pedestrian detection windows can be quickly rejected by the first few stages, it accelerates detection speed by 1.43 times. By eliminating the highly overlapped detection windows with lower scores after the first stage, it's 4.07 times faster with negligible performance loss.
\end{abstract}

\begin{IEEEkeywords}
Pedestrian Detection, Multilayer Channel Features (MCF), HOG+LUV, CNN, NMS.
\end{IEEEkeywords}

%
\IEEEpeerreviewmaketitle

\section{Introduction}
%
%
%
%
\IEEEPARstart{P}{edestrian} detection based on Convolutional Neural Network (i.e., CNN) has achieved great success recently \cite{Sermanet_UMFL_CVPR_2013}, \cite{Yang_CCF_ICCV_2015}, \cite{Hosang_Deeper_CVPR_2015}, \cite{Li_ScaleAware_arXiv_2015}, \cite{Benenson_TenYears_ECCV_2014}. The main process of CNN based methods can be divided into two steps: proposal extraction and CNN classification. Firstly, the candidate proposals are extracted by the traditional pedestrian detection algorithm (e.g., ACF \cite{Dollar_ACF_PAMI_2014} and LDCF \cite{Nam_LDCF_NIPS_2014}). Then, these proposals are classified into pedestrian or non-pedestrian by the CNN \cite{Hosang_Deeper_CVPR_2015}, \cite{Cai_CompACT_ICCV_2015}.
	
Despite its great success, it still exists some room to improve it. 1) Most methods only use the last layer features in CNN with softmax or SVM to classify the proposals. In fact, different layers in CNN represents the different image characteristic. The first few layers can better describe the image local variance, whereas the last few layers abstract the image global structure. It means that each layer in CNN contains different discriminative features, which can be used for learning the classifier. 2) Some methods only use the traditional methods based on the handcrafted features (i.e., HOG+LUV \cite{Dollar_ICF_BMVC_2009}) to generate the candidate proposals while ignoring the proposal scores. 3) Due to the large amount of convolutional operations, the methods based very deep CNN (e.g., VGG16 \cite{Simonyan_VGG_arXiv_2015}) run very slowly on the common CPU (e.g., about 8s).

Recently, researchers have done some work to solve the above problems. Li et al. \cite{Li_CNNC_CVPR_2015} proposed to train the cascaded multiple CNN models of different resolutions. As the low resolution CNN can early reject many background regions, it avoids scanning the full image with high resolution CNN and then reduces the computation cost. However, the training process of multiple CNN models is relatively complex. Cai et al. \cite{Cai_CompACT_ICCV_2015} proposed the complexity-aware cascade to seamlessly integrate handcrafted features and the last layer features in CNN into a unifying detector. However, it still does not make full use of the multi-layer features in CNN. Bell et al. \cite{Bell_IONet_arXiv_2015} concatenated the multiple layers of CNN into the fixed-size ROI pooling. With more abundant feature abstraction, it outperforms fast-RCNN \cite{Girshick_Fast_ICCV_2015}. Though its success, it needs complex operations of L2-normalized, concatenated, scaled, and dimension-reduced. Moreover, it ignores the scores of the proposals.

In this paper, we propose a unifying framework, which is called Multi-layer Channel Features (MCF). Firstly, it integrates handcrafted image channels (i.e., HOG+LUV) and each layer of CNN into the multi-layer image channels. HOG+LUV image channels are set as the first layer, which contains 10 image channels. The layers in CNN correspond to the remaining layers, respectively. Secondly, zero-order, one-order, and high-order features are extracted to generate a large number of candidate feature pools in each layer. Finally, a multi-stage cascade AdaBoost is used to select the discriminative features and efficiently classify object and background. The weak classifiers in each stage of multi-stage cascade are learned based on the candidate features from corresponding layer. To further accelerate detection speed, the highly overlapped detection windows with lower scores are eliminated after the first stage. Overall, the contributions of this paper and the merits of the proposed methods (MCF) can be summarized as follows:

\begin{enumerate}
\item The unifying framework MCF is proposed. MCF seamlessly integrates HOG+LUV image channels and each layer of CNN into a unifying multi-layer image channels. Due to the diverse characteristic of different layers, these layers can provide more rich feature abstractions.

\item  Multi-stage cascade AdaBoost is learned from multi-layer image channels. It can achieve better performance with more abundant feature abstractions and quickly reject many detection windows by the first few stages.

\item  The highly overlapped detection windows with lower scores are eliminated after the first stage. Thus, it can reduce the large computation cost of CNN operations. With very little performance loss, it's 4.07 times faster. Finally, it's possible that MCF with very deep CNN (e.g., VGG16 \cite{Simonyan_VGG_arXiv_2015}) can run at 0.54 fps on the common CPU, while it achieves 11.05\% miss rate on original Caltech pedestrian set.

\item  MCF achieves the state-of-the-art performance on Caltech pedestrian dataset (the log-average miss rate is 10.40\%), which outperforms CompACT-Deep \cite{Cai_CompACT_ICCV_2015} by 1.35\%. Using new and more accurate annotations \cite{Zhang_RotatedFilters_arXiv_2016} of the test set, MCF achieves 7.98\% miss rate, which is superior to other methods.

\end{enumerate}

The rest of the paper is organized as follows. Firstly, we give a review about pedestrian detection. Then, our methods are introduced in Sec. III. Sec. IV shows the experimental results. Finally, we conclude this paper in Sec. V.

\section{Related Work}
According to whether or not CNN is used, pedestrian detection can be divided into two main manners: the handcrafted channels based methods and CNN based methods. Handcrafted channels based methods are relatively simple and efficient, whereas CNN based methods are much more effective but inefficient. We firstly give a review about the handcrafted channels based methods and then introduce some methods based on CNN.

Haar features based cascade AdaBoost detector is one of the most famous object detection methods \cite{Viola_Haar_IJCV_2004}, \cite{Pang_SF_TC_2016}. It can quickly reject a large number of non-object detection windows by the early stages of the cascade. Dalal and Triggs \cite{Dalal_HOG_CVPR_2005} proposed to use the Histogram of Oriented Gradients (HOG) to describe the image local variance. It can work very well with a linear SVM. To handle pose variations of objects, Felzenszwalb et al. \cite{Felzenszwalb_DPM_PAMI_2010} proposed the Deformable Part Model (DPM) based on HOG features, which is a mixture of six deformable part models and one root model.

By integrating cascade AdaBoost \cite{Viola_Haar_IJCV_2004}, \cite{Bourdev_SoftCascade_CVPR_2005} and HOG features \cite{Dalal_HOG_CVPR_2005}, Doll\'ar et al. \cite{Dollar_ICF_BMVC_2009} proposed Integral Channel Features (ICF). Firstly, it extracts the local sum features from HOG channels and LUV color channels (i.e., HOG+LUV). Then, cascade AdaBoost \cite{Bourdev_SoftCascade_CVPR_2005}, \cite{Zhang_MIP_NIPS_2007} is used to learn the classifier. To further speedup the detection, Doll\'ar et al. \cite{Dollar_ACF_PAMI_2014} then proposed Aggregated Channel Features (ACF), which downsamples the image channels by a factor of 4. 

Following ICF \cite{Dollar_ICF_BMVC_2009}, SquaresChnFtrs \cite{Benenson_SquaresFtrs_CVPR_2013}, InformedHaar \cite{Zhang_InformedHaar_CVPR_2014}, LDCF \cite{Nam_LDCF_NIPS_2014}, Filtered Channel Features (FCF) \cite{Zhang_FCF_CVPR_2015}, and NNNF \cite{Cao_NNNF_arXiv_2015} have also been proposed. They all employ the same image channels (i.e., HOG+LUV) as ICF. In SquaresChnFtrs \cite{Benenson_SquaresFtrs_CVPR_2013}, the pixel sums of local square regions in each channel are used for learning the classifier. InformedHaar \cite{Zhang_InformedHaar_CVPR_2014} incorporates the statistical pedestrian model into the design of simple haar-like features. Inspired by \cite{Hariharan_DD_ECCV_2012}, Nam et al. \cite{Nam_LDCF_NIPS_2014} proposed to calculate the decorrelated channels by convolving the PCA-like \cite{Pang_LDA_TNNLS_2014} filters with HOG+LUV image channels. Recently, Zhang et al. \cite{Zhang_FCF_CVPR_2015} proposed to put the above different types of channel features into a unifying framework (i.e., FCF). FCF generates the candidate feature pool by convolving a filter bank (RadomFilters, Checkerboards, etc) with HOG+LUV image channels. It's found that using the simple Checkboards filters could achieve very good performance. Based on the appearance constancy and shape symmetry, Cao et al. \cite{Cao_NNNF_arXiv_2015} proposed NNNF features.

Recently, deep Convolutional Neural Network (CNN) based methods have also achieved great success in object detection \cite{Krizhevsky_AlexNet_NIPS_2012}, \cite{Girshick_RCNN_CVPR_2014}, \cite{Bell_IONet_arXiv_2015}, \cite{Ren_Faster_NIPS_2015}, \cite{Girshick_Fast_ICCV_2015}, \cite{Zhang_Cosaliency_TNNLS_2015}, \cite{Li_Lane_TNNLS_2016}. Generally speaking, it firstly generates the candidate object proposals \cite{Cheng_BING_CVPR_2014}, \cite{Uijlings_SS_IJCV_2013} and then uses the trained CNN model \cite{Krizhevsky_AlexNet_NIPS_2012}, \cite{Girshick_RCNN_CVPR_2014} to classify these proposals. Hosang et al. \cite{Hosang_Deeper_CVPR_2015} generalized CNN model for pedestrian detection after using the handcrafted features based methods to extract the candidate pedestrian proposals. To eliminate the number of hard negative proposals in the background, Tian et al. \cite{Tian_Tian_CVPR_2015} proposed to jointly optimize pedestrian detection with semantic tasks. Recently, Tian et al. \cite{Tian_DeepParts_ICCV_2015} proposed to learn deep strong part models to handle the problem of pedestrian occlusion. Li et al. \cite{Li_ScaleAware_arXiv_2015} proposed the scale-aware fast-RCNN by incorporating a large scale sub-network and a small scale sub-network into a unifying architecture.

Despite the success of CNN based pedestrian detection, it still exists some room to improve it. Firstly, the score information of the candidate proposals can be used to boost the detection performance. Secondly, each layer in CNN contains some discriminative features, which can be used for learning classifier and rejecting non-pedestrian detection windows early. Cai et al. \cite{Cai_CompACT_ICCV_2015} proposed to seamlessly integrate CNN and handcrafted features. Though it uses the proposal score information, it still ignores features of the inner layer of CNN. Sermanet et al. \cite{Sermanet_UMFL_CVPR_2013} proposed to concatenate the first layer feature and the second layer together. Bell et al. \cite{Bell_IONet_arXiv_2015} proposed to use skip pooling to integrate multiple layers. It's called skip-layer connections. Despite its success, there is still some problems: 1) It ignores the proposals scores; 2) The proposal should pass through the whole CNN before classification; 3) The skip-layer operations in \cite{Bell_IONet_arXiv_2015} is relatively complex.

\section{Our Methods}
\subsection{Multi-layer Channel Features (MCF)}
The layers in CNN represent the different and diverse image characteristic. The image channels in the first few layers can better describe the image local variance, while the image channels in the last few layers can abstract the image global structure. Meanwhile, the handcrafted image channels (e.g., HOG+LUV) can be also able to describe the image variations very well. HOG channels can describe the image local edge directions and variances. LUV channels capture the image color information. Compared to the layers in CNN, the handcrafted image channels are very simple and efficient. In this paper, we integrate HOG+LUV and the layers of CNN to construct Multi-layer Channel Features (MCF).

\begin{figure}[!t]
\label{MCF}
\centering
\includegraphics[width=3.2in]{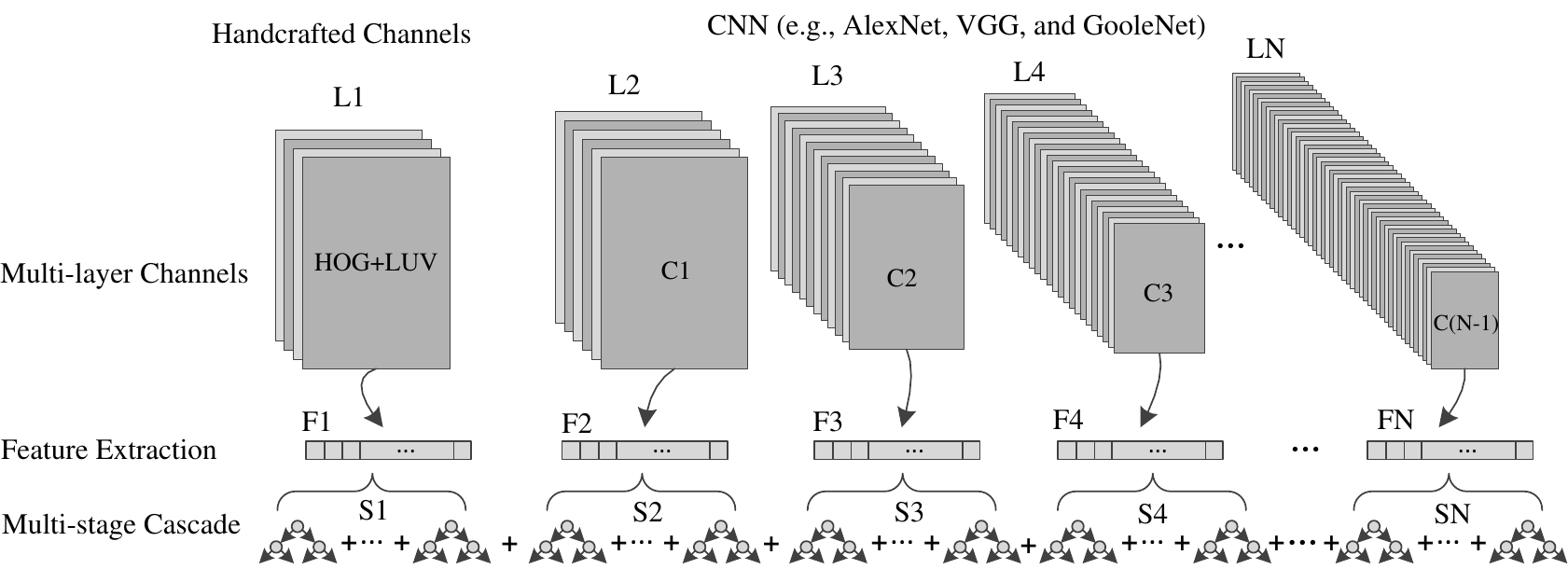}
\caption{The basic architecture of MCF. It can be divided into three steps: multi-layer channel generation, feature extraction, and multi-stage cascade AdaBoost classifier.} 
\end{figure}

First of all, we give an overview about our proposed Multi-layer Channel Features (i.e., MCF). Fig. 1 show the basic architecture of MCF. It can be divided into three parts: 1) Firstly, multi-layer image channels from L1 to LN are generated. The traditional handcrafted image channels (i.e., HOG+LUV) are used for the first layer (i.e., L1). The convolutional layers from C1 to C(N-1) in CNN construct the remaining layers from L2 to LN. In each layer, there are multiple image channels. 2) The second step is feature extraction. Zero-order, one-order, and high-order features can be calculated in the image channels of each layer. 3) Finally, the multi-stage cascade AdaBoost is learned from the candidate features of each layer one after another. The weak classifiers in each stage of multi-stage cascade are learned from the candidate features of corresponding layer. For example, the weak classifiers in Stage 2 (i.e., S2) are learned from candidate features F2 of Layer 2 (i.e., L2).

Fig. 2 shows the test process of MCF. Give the test image, the image channels in L1 (i.e., HOG+LUV) are firstly computed. Detection windows are generated by scanning the test image. These detection windows are classified by S1 using the weak classifiers learned from L1. Some detection windows will be rejected by S1. For the detection windows accepted by S1, the image channels in L2 are computed. Then the accepted detection windows are classified by S2 using the weak classifiers learned from L2. The above process is repeated from L1 to LN. Finally, the detection windows accepted by all the stages (i.e., S1 to SN) will be merged by NMS. The merged detection windows are the final pedestrian windows.

\begin{figure}[!t]
\label{TestMCF}
\centering
\includegraphics[]{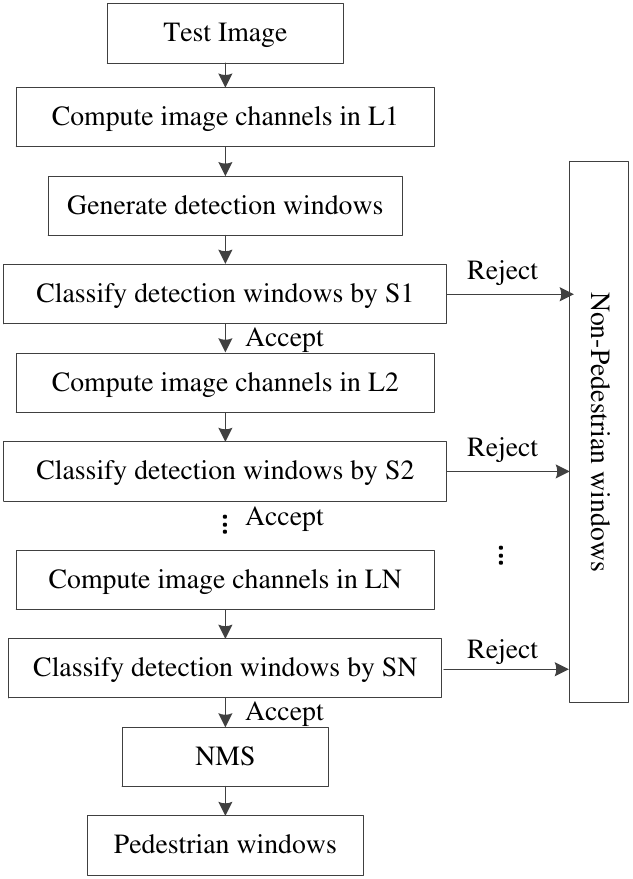}
\caption{Test process of basic MCF.} 
\end{figure}

\textbf{Multi-layer Image Channels} Row 1 in Fig. 1 shows the multi-layer image Channels. It consists of \textit{N} layers. In each layer, there are multiple image channels. Table \ref{TabParam} shows the specific parameters of multi-layer image channels based on HOG+LUV and VGG16. It contains six layers from L1 to L6. L1 is the handcrafted image channels (i.e., HOG+LUV). L2-L6 are five convolutional layers (i.e., C1-C5) in VGG16. Row 3 shows the image size in each layer. The image size in L1 is $128\times64$. The sizes of L2-L6 are $64\times32$, $32\times16$, $16\times8$, $8\times4$, and $4\times2$, respectively. Row 4 shows the number of the channels in each layer. L1 contains 10 image channels. L2-L6 each have 64, 128, 256, 512, and 512 image channels, respectively. In Table \ref{TabParam}, all the convolutional layers in CNN (i.e., C1 to C5) are used for constructing the multi-layer image channels. In fact, only part convolutional layers in CNN can also construct the multi-layer image channels. For example, a five-layer image channels can be generated by HOG+LUV and C2-C5 of VGG16. C1 of VGG16 is not used.  The corresponding MCF are shown in Fig. 3.

\begin{figure}[!t]
\label{MCF_L5}
\centering
\includegraphics[width=3in]{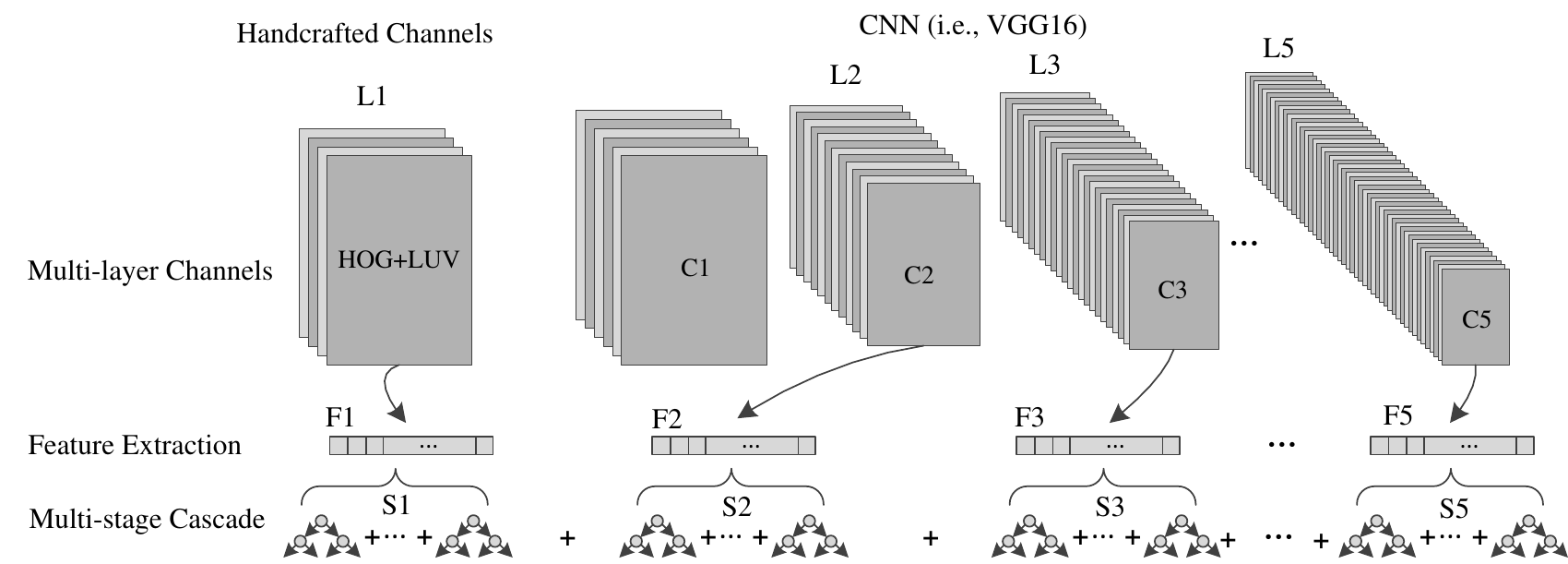}
\caption{MCF are generated by HOG+LUV and C2-C5 of VGG16. C1 of VGG16 are not used for constructing MCF.} 
\end{figure}

\begin{table}[!t]
\centering
\renewcommand{\arraystretch}{1.3}
\caption{Multi-layer image channels. The first layer is HOG+LUV, the remaining layers are the convolutional layers (i.e., C1 to C5) in VGG16. }
\begin{tabular}
{|c|c|c|c|c|c|c|c|}
\hline
Layer& L1& L2& L3& L4& L5& L6\\
\hline
\multirow{2}*{Name} & HOG& \multicolumn{5}{c|}{VGG16}  \\
\cline{3-7}
 & LUV & C1 & C2& C3& C4& C5 \\
\hline
Size& $128\times64$& $64\times32$& $32\times16$& $16\times8$& $8\times4$& $4\times2$ \\
\hline
Num& 10& 64& 128& 256& 512& 512\\
\hline
\end{tabular}
\label{TabParam}
\end{table}

\textbf{Feature Extraction} Features can be divided into three classes: zero-order feature, one-order feature, and high-order feature. In zero-order feature extraction, a single pixel itself is used as a feature and no neighboring pixels are used. One-order feature is defined as the pixel sums or averages in the local or non-local regions in each channel.  High-order feature defined by the difference of the sums or averages of two or more different regions. 
For L1 (i.e., HOG+LUV), there are many successful methods for feature extraction, including ICF \cite{Dollar_ICF_BMVC_2009}, ACF \cite{Dollar_ACF_PAMI_2014}, SquaresChnFtrs \cite{Benenson_SquaresFtrs_CVPR_2013}, InformedHaar \cite{Zhang_InformedHaar_CVPR_2014}, LDCF \cite{Nam_LDCF_NIPS_2014}, FCF \cite{Zhang_FCF_CVPR_2015}, and NNNF \cite{Cao_NNNF_arXiv_2015}. ICF, ACF, and SquaresChnFtrs can be seen as one-order features. InformedHaar, LDCF, FCF, and NNNF are high-order features. Compared to the other features, ACF has the fastest detection speed. NNNF has the best trade-off between detection speed and detection performance. Due to the simplicity and effectiveness, ACF and NNNF are used for feature exaction in L1.  The number of image channels from CNN is relatively large. For example, the fourth convolutional layer (i.e., C4) in VGG16 has 512 image channels (see Table \ref{TabParam}). To reduce the computation cost and avoid a very large number of candidate features, only zero-order feature is used. It means that each pixel value in image channels  of each layer is used for the candidate feature. The specific feature extraction in multi-layer image channels can be seen in Fig. 4.

\begin{figure}[!t]
\label{FeaExtraction}
\centering
\includegraphics[width=3.2in]{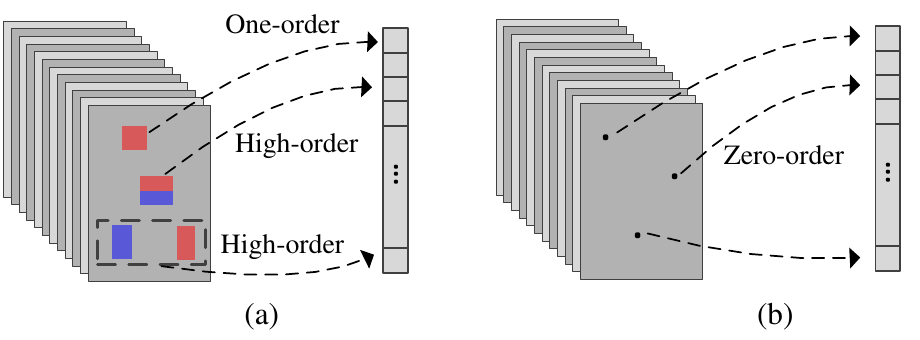}
\caption{Feature Extraction in Multi-layer image channels. (a) feature extraction in L1 (HOG+LUV), where one-order (ACF) and high-order features (NNNF) are used. (b) feature extraction in L2-LN (the layers of CNN), where zero-order features are extracted. Zero-order feature means that a single pixel value in each channel is used as a feature. 
} 
\end{figure}

\textbf{Multi-stage Cascade AdaBoost} Cascade AdaBoost is a popular method for object detection. Based on multi-layer image channels, we propose the multi-stage cascade AdaBoost for pedestrian detection. Fig. 5 gives the specific explanations about multi-stage cascade. The features in Si are learned from the candidate features Fi  of Li, where i=1, 2, ..., N. Firstly, $k_1$ weak classifiers in S1 are learned from the candidate features F1 extracted from L1. Based on the hard negative samples and positive samples, $k_2$ weak classifiers in S2 are then learned from F2. The remaining stages are trained in the same manner. Finally, multi-stage (i.e., N-stage) cascade AdaBoost classifier can be obtained. This strong classifier $H(\bf{x})$ can be expressed as the following equation:
\begin{equation}
\begin{split}
H(\mathbf{x})&=\sum\limits_{j=1}^{k_1}{h_1^j(\mathbf{x})}+...+\sum\limits_{j=1}^{k_i}{h_i^j(\mathbf{x})}+...+\sum\limits_{j=1}^{k_N}{h_N^j(\mathbf{x})}
\\&=\sum\limits_{i=1}^{N}\sum\limits_{j=1}^{k_i}{h_i^j(\mathbf{x})},
\end{split}
\end{equation}
where $\bf{x}$ represents the samples (windows), $h_i^j(\bf{x})$ represents the $j$-th weak classifier in Stage $i$. $k_1$, $k_2$, ..., $k_N$ are the number of weak classifiers in each stage, respectively. How to set the value of $k_1$, $k_2$, ..., $k_N$  is an open problem. In this paper, one of simple structure is used  as the follows: 
\begin{equation}
\begin{split}
& k_1=N_{All}/2,
\\& k_2=k_3=...=k_N=N_{All}/(2\times(N-1)),
\end{split}
\end{equation}
where $N_{All}$ represents the number of the total weak classifiers. Based on soft-cascade \cite{Bourdev_SoftCascade_CVPR_2005}, the reject thresholds are set after each weak classifier.

The advantages about the multi-stage cascade AdaBoost structure can be concluded as the following: 1) Firstly, it avoids learning classifier from a very large feature pooling (e.g., more than one million); 2) Secondly, it makes full use of the information from multi-layer image channels. Thus, it can enrich the feature abstraction. 3) Finally, many non-pedestrian detection windows can be quickly rejected by the features in the first few layers. Thus, it avoids the computation cost of the remaining layers in CNN and accelerates the detection speed.

\begin{figure}[!t]
\label{MCA}
\centering
\includegraphics[width=3.2in]{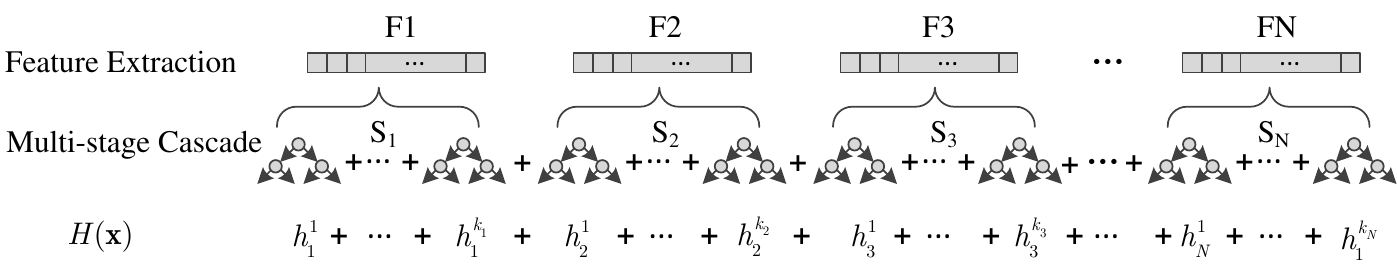}
\caption{Multi-stage cascade AdaBoost. Each stage learns the classifiers from the candidate features of corresponding layer.} 
\end{figure}

\subsection{Elimination of Highly Overlapped Windows}
Pedestrian detection is a multiple instance problem. Generally, the adjacent area around the pedestrian exists many positive detection windows. Many of these positive detection windows around pedestrians highly overlap. Though multi-stage cascade AdaBoost structure can reject many non-pedestrian detection windows, it cannot reject the positive detection windows around pedestrians. When the cascade classifier based on very deep CNN (e.g., VGG16), the computation cost of these positive detection windows are large.

In fact, there is no need to put all the highly overlapped windows accepted by first stage into the remaining stages. Detection windows accepted by the first stage each have a classification score. The highly overlapped windows with lower scores can be eliminated after the first stage. To eliminate these highly overlapped windows with lower scores, Non-Maximum Suppression (i.e., NMS) is used after the first stage. The overlap ratio $O(w_1,w_2)$ of detection windows can be defined in the following:
\begin{equation}
O(w_1,w_2)=\frac{area(w_1 \cap w_2)}{area(w_1 \cup w_2)},
\end{equation}
where $w_1$ and $w_2$  are two detection windows. If $O(w_1,w_2)>\theta$, it means that $w_1$ and $w_2$ highly overlap. Then the detection window with lower score will be eliminated. Instead of the standard threshold $\theta=0.5$, a larger threshold is used here. Experimental results show that $\theta=0.8$ can accelerate the detection speed with little performance loss. Fig. 6 shows the specific test process of MCF by eliminating highly overlapped detection windows.

\begin{figure}[!t]
\label{MCF-NMS}
\centering
\includegraphics[width=3.2in]{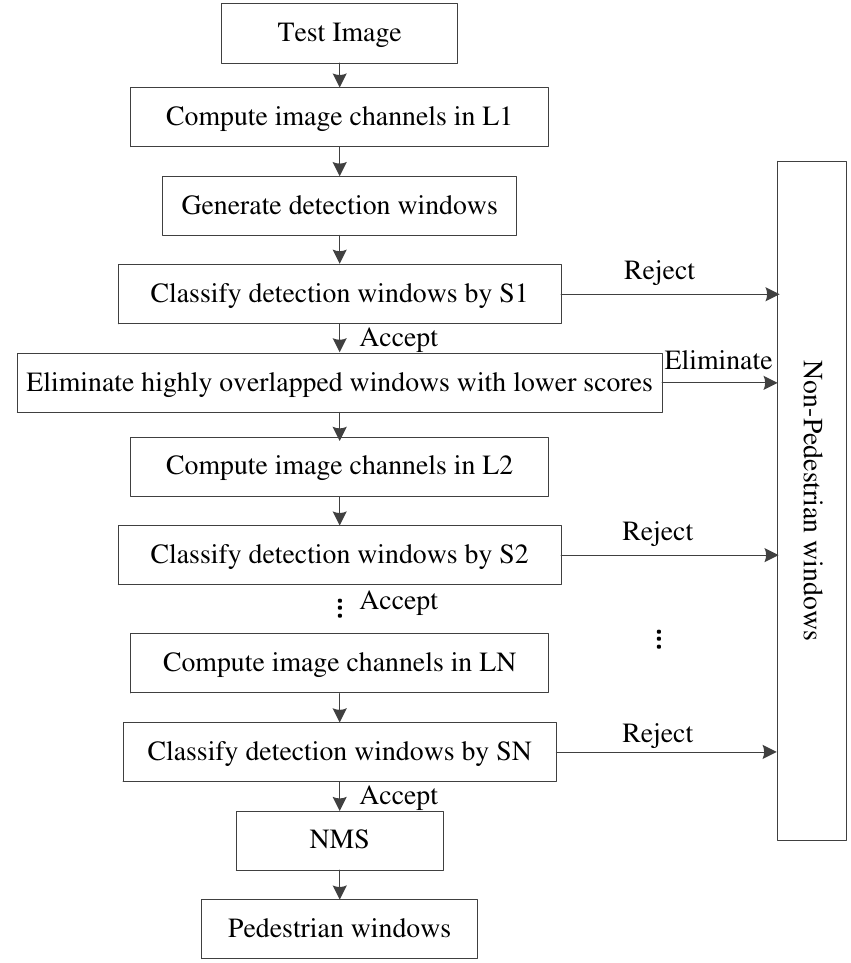}
\caption{Test process of the fast version of MCF where the technique of NMS is used to eliminating highly overlapped detection windows with lower scores.
} 
\end{figure}

\section{Experiments}
The challenging Caltech pedestrian detection dataset \cite{Dollar_PD_CVPR_2009}, \cite{Dollar_PD_PAMI_2010} is employed for the evaluation. It consists of 11 videos. The first 6 videos are used for training and the rest videos are used for testing. The raw training images are formed by sampling one image per 30 frames. It results in 4250 images for training, where there are 1631 positive samples. The corresponding training data is called Caltech.

To enlarge the training samples, Caltech10x is used. It samples one image per 3 frames in the training videos. As a result, there are 42,782 images in which there are 16,376 positive samples. Please note that the testing data is same as \cite{Dollar_PD_CVPR_2009}, \cite{Dollar_PD_PAMI_2010} whenever the Caltech or Caltech10x is used. It contains 4024 images in which there are 1014 pedestrians.

The first layer in MCF is HOG+LUV image channels \cite{Dollar_ICF_BMVC_2009}, which contains one normalized gradient magnitude channel, six histograms of oriented gradient channels, and three color channels. Two popular CNN models (i.e., AlexNet \cite{Krizhevsky_AlexNet_NIPS_2012} and VGG16 \cite{Simonyan_VGG_arXiv_2015}) are used for constructing the remaining layers in MCF. Instead of using original input size $227\times227$ or $224\times224$, we use the size $128\times64$ for pedestrian detection. For AlexNet, stride 4 in the first convolutional layer is replaced by stride 2. The input size $6\times6$ in the first full connected layer is replaced by the size $8\times4$. For VGG16, the input size $7\times7$ of the first full connected layer is replaced by the size $4\times2$. The other initial parameters follow the pre-trained models on ImageNet. The final parameters in AlexNet and VGG16 are fine-tuned on the Caltech10x.

Feature extraction in L1 (i.e., HOG+LUV) is ACF \cite{Dollar_ACF_PAMI_2014} (Section IV.A) or NNNF \cite{Cao_NNNF_arXiv_2015} (Section IV.B). Feature extraction in the remaining layers (i.e., the layers of CNN) is zero-order feature (single pixel). The final classifier consists of 4096 level-2 or level-4 decision trees. The decision tree number of each stage are $k_1=2048$, $k_2=k_3=...=k_N=2048/(N-1)$, respectively. $N$ is the number of the layers in MCF.

\subsection{Self-Comparison of MCF}
In this section, some intermediate experimental results on original Caltech training set are reported to show how to setup the effective and efficient MCF. Some specific experimental setup is as follows. HOG+LUV are used for the first layer. The convolutional layers in CNN (i.e., AlexNet or VGG16) correspond to the remaining layers. The feature extraction in HOG+LUV is ACF. Feature extraction in the layers of CNN is just zero-order feature (single pixel). To speed up the training, negative samples are generated by five round training of original ACF \cite{Dollar_ACF_PAMI_2014}, where the number of trees in each round are 32, 128, 512, 1024, and 2048, respectively. Finally, multi-stage cascade which consists of 4096 level-2 decision trees is learned based on these negative samples and positive samples. The first stage contains the first 2048 decision trees. The remaining stages equally split the remaining 2048 decision trees. For example, HOG+LUV and C2 to C5 of CNN construct a five-layer image channels. Then, the corresponding five-stage cascade can be learned. The first stage S1 has 2048 weak classifiers. The remaining stages (i.e., S2-S5) each have 512 weak classifiers. Miss Rates are log-averaged over the range of FPPI = [$10^{-2}$,$10^{0}$], where FPPI represents False Positive Per Image. 

\begin{table}[!t]
\centering
\renewcommand{\arraystretch}{1.4}
\caption{Miss Rates (MR) of MCF based on HOG+LUV and the different layers in CNN. $\surd$ means that the corresponding layer is used. HOG+LUV is always used for the first layer. The layers in AlexNet or VGG16 are used for the remaining layers.}
\begin{tabular*}{8.5cm}{@{\extracolsep{\fill}}ccccccccc}
\hline
\multirow{2}*{Name} & HOG & \multicolumn{5}{c}{AlexNet} & \multirow{2}*{MR (\%)} & \multirow{2}*{$\Delta$ MR (\%)}  \\
\cline{3-7}
 & LUV & C1 & C2 & C3& C4& C5 & & \\
\hline
MCF-2 & $\surd$& & & & & $\surd$& 20.08& N/A \\
MCF-3 & $\surd$& & & & $\surd$& $\surd$& 18.43& 1.65\\
MCF-4 & $\surd$& & & $\surd$& $\surd$& $\surd$& 17.40& 2.68\\
MCF-5 & $\surd$& & $\surd$& $\surd$& $\surd$& $\surd$& 18.01& 2.07\\
MCF-6 & $\surd$& $\surd$& $\surd$& $\surd$& $\surd$& $\surd$& 17.29& 2.79\\
\hline
\hline
\multirow{2}*{Name} & HOG & \multicolumn{5}{c}{VGG16} & \multirow{2}*{MR (\%)} & \multirow{2}*{$\Delta$ MR (\%)}  \\
\cline{3-7}
 & LUV & C1 & C2 & C3& C4& C5 & & \\
\hline
MCF-2 & $\surd$& & & & & $\surd$& 18.52& N/A \\
MCF-3 & $\surd$& & & & $\surd$& $\surd$& 17.14& 1.38\\
MCF-4 & $\surd$& & & $\surd$& $\surd$& $\surd$& 15.40& 3.12\\
MCF-5 & $\surd$& & $\surd$& $\surd$& $\surd$& $\surd$& 14.78& 3.74\\
MCF-6 & $\surd$& $\surd$& $\surd$& $\surd$& $\surd$& $\surd$& 14.31& 4.21\\
\hline
\end{tabular*}
\label{TabMiss}
\end{table}

Table \ref{TabMiss} shows Miss Rates (MR) of MCF based on HOG+LUV and the different layers in CNN. The results based on AlexNet and VGG16 are both shown here. $\surd$ means that the corresponding layer is used for MCF. HOG+LUV image channels are always used for the first layer. The layers (i.e., C1, C2, ... or C5) are used for the remaining layers. MCF-N means that there are N layers in MCF. For example, MCF-3 in Row 3 are generated by HOG+LUV, C4 and C5 of AlexNet. The first layer is HOG+LUV image channels. The second layer is the fourth convolutional layer (i.e., C4) of AlexNet. The last layer is the fifth layer (i.e., C5) of AlexNet. Based on multi-layer image channels, the corresponding multi-stage cascade is learned. There are the following observations from Table \ref{TabMiss}: 1) Compared to MCF-2, MCF-N (N$>$2) usually achieves the better performance. For example, the miss rate of MCF-6 based on VGG16 in the last row is lower than that of MCF-2 by 4.21\%; 2) Generally, with increase of the layer number, the miss rate of MCF becomes lower and the detection performance becomes better. The above observations demonstrate that the middle layers in CNN can enrich the feature abstraction. It means that each layer in CNN contains some discriminative features, which can be used for classification.

\begin{table}[!t]
\centering
\renewcommand{\arraystretch}{1.3}
\caption{Rejected number and rejected ratio by the stages in MCF-6 are shown. `*' means that the average number of detection windows accepted by stage 1 are shown.}
\begin{tabular*}{8.5cm}{@{\extracolsep{\fill}}ccccc}
\hline
\multirow{2}*{Stage}  & \multicolumn{2}{c}{HOG+LUV and AlexNet} & \multicolumn{2}{c}{HOG+LUV and VGG16}  \\
\cline{2-3} \cline{4-5}
 & Number& Ratio & Number& Ratio\\
\hline
S1 & 159*& N/A& 159* & N/A \\
\hline
S2 & 35& 22.0\%& 23& 14.5\%\\
S3& 35& 22.0\%& 21 & 13.2\%\\
S4& 21& 13.2\%& 33& 20.8\%\\
S5& 14& 8.8\%& 29& 18.2\%\\
S6& 8& 5.0\%& 15& 9.4\%\\
Total& 113& 71.0\%& 121& 76.1\%\\
\hline
\end{tabular*}
\label{TabReject}
\end{table}

Table \ref{TabReject} shows the average number and the ratio of detection windows rejected by each stage in MCF-6. MCF-6 is based on HOG+LUV and all the five convolutional layers in CNN. Thus, the multi-stage cascade AdaBoost in MCF has six stages from S1 to S6. `*' means that the average number of detection windows accepted by stage 1, instead of that rejected by stage 1, is shown. As the weak classifiers in S1 are both learned from HOG+LUV, the number of detection windows accepted by S1 are same (i.e., 159). Among the 159 accepted detection windows, about 71\% and 76.1\% detection windows are rejected by the cascade based on AlexNet and that based on VGG16, respectively. Overall, the multi-stage cascade based on VGG16 can reject more detection windows. Specifically, the first two stage stages (i.e., S2 and S3) based on AlexNet reject more detection windows than that based on VGG16. The middle two stages (i.e., S4 and S5) based on VGG16 can reject more detection windows than that based on AlexNet.

\begin{table}[!t]
\centering
\renewcommand{\arraystretch}{1.3}
\caption{Miss Rates (MR) and detection time of MCF-2 and MCF-6. MCF-2 is based on HOG+LUV and C5 of CNN. MCF-6 is based on HOG+LUV and C1-C5 of CNN.}
\begin{tabular*}{8.5cm}{@{\extracolsep{\fill}}ccccc}
\hline
 & \multicolumn{2}{c}{HOG+LUV and AlexNet} & \multicolumn{2}{c}{HOG+LUV and VGG16}  \\
\cline{2-3} \cline{4-5}
 & MCF-2& MCF-6& MCF-2& MCF-6\\
\hline
MR (\%)& 20.08& 17.29& 18.52& 14.31\\
Time (s)& 2.99& 2.30& 7.69& 5.37\\
\hline
\end{tabular*}
\label{TabSum1}
\end{table}

As multi-stage cascade can reject many detection windows by the first few stages, MCF can accelerate the detection speed. Table \ref{TabSum1} compares the detection time and detection performance between MCF-2 and MCF-6. MCF-2 uses HOG+LUV and C5 in CNN to construct two-layer image channels. Then two-stage cascade is learned. MCF-6 uses HOG+LUV and all the five convolutional layers from C1 to C5 in CNN to construct six-layer image channels. Then six-stage cascade is learned. The detection time shown in Table \ref{TabSum1}  is based on the common CPU (i.e., Intel Core i7-3700). No matter CNN model is AlexNet or VGG16, MCF-6 have the better performance and the faster detection speed. For example, based on VGG16, the miss rates of MCF-2 and MCF-6 are 18.52\% and 14.31\%, respectively. The detection times of MCF-2 and MCF-6 are 7.69s and 5.37s, respectively. Thus, the miss rate of MCF-6 is lower than that of MCF-2 by 4.21\%, while the speed of MCF-6 is 1.43 times faster than that of MCF-2. The reasons can be explained as the following: 1) As MCF-6 uses all the layers in CNN to learn the classifier, it can learn more abundant features. Thus, it has a better performance. 2) MCF-2 needs to calculate all the layers of CNN (i.e., C1 to C5) before classifying the detection windows accepted by S1. MCF-6 just needs to calculate the $i$-th layer of CNN before classifying the detection windows by Si (i=2,3,...,6). In the Table \ref{TabReject}, MCF-6 reject 66.7\% detection windows before S6. Thus, MCF-6 has faster detection speed than MCF-2.

\begin{table}[!t]
\centering
\renewcommand{\arraystretch}{1.3}
\caption{Miss rates and detection time vary with $\theta$. MCF used here is based on HOG+LUV and AlexNet.}
\begin{tabular*}{8.5cm}{@{\extracolsep{\fill}}ccccc}
\hline
\multirow{2}*{$\theta$}& \multicolumn{2}{c}{MCF-2} & \multicolumn{2}{c}{MCF-6}  \\
\cline{2-3} \cline{4-5}
 & MR (\%)& Time (s)& MR (\%)& Time (s)\\
\hline
INF& 20.08& 2.99& 17.29& 2.30\\
0.50& 23.70& 0.44& 21.65& 0.34\\
0.80& 20.97& 1.15& 18.06& 0.86\\
0.85& 20.30& 1.76& 17.34& 1.35\\
0.90& 19.82& 2.03& 17.32& 1.57\\
\hline
\end{tabular*}
\label{TabNMS}
\end{table}

Though the speed of MCF-6 is faster than that of MCF-2, it's still very slow. To further accelerate the detection speed, the highly overlapped detection windows with lower scores accepted by the first stage (i.e., S1) are eliminated by NMS. As stated in section III.B, the threshold $\theta$ is an important factor to balance detection speed and detection performance. Table \ref{TabNMS} shows that miss rates and detection time vary with $\theta$. MCF-2 and MCF-6 based on HOG+LUV and AlexNet in Table \ref{TabSum1} are used for the baseline (i.e., $\theta=$INF). When $ \theta =0.5$, the detection speed is very fast, but the detection performance drops rapidly. For MCF-6, the detection speed of MCF-6 with $\theta=0.5$ is 6.76 times faster than original MCF, while the miss rate of MCF with $\theta=0.5$ is higher than original MCF by 3.36\%. Thus, it's not a good choice. When  $\theta=0.9$, the detection performance is almost no loss, while the detection speed is not significantly improved. Thus, the trade-off choice is $\theta=0.8$. With little performance loss (e.g., 0.77\%) in MCF-6, it's 2.67 times faster than original MCF. In the following section, MCF with $\theta=0.8$  are called MCF-f.

\begin{table}[!t]
\centering
\renewcommand{\arraystretch}{1.3}
\caption{Miss Rate (MR) and detection time of MCF-2, MCF-6, and MCF-6-f. MCF-2 is based on HOG+LUV and C5 in CNN. MCF-6 is based on HOG+LUV and C1-C5 in CNN. MCF-6-f is the fast version of MCF-6.}
\begin{tabular*}{8.5cm}{@{\extracolsep{\fill}}cccc}
\hline
 & \multicolumn{3}{c}{HOG+LUV and AlexNet}\\
\cline{2-4}
 & MCF-2& MCF-6& MCF-6-f\\
\hline
MR (\%)& 20.08& 17.29& 18.06\\
Time (s)& 2.99& 2.30& 0.86\\
\hline
\hline
 & \multicolumn{3}{c}{HOG+LUV and VGG16}\\
\cline{2-4}
 & MCF-2& MCF-6& MCF-6-f\\
\hline
MR (\%)& 18.52& 14.31& 14.89\\
Time (s)& 7.69& 5.37& 1.89\\
\hline
\end{tabular*}
\label{TabSum2}
\end{table}

Table \ref{TabSum2} summarizes MCF-2, MCF-6 and MCF-6-f. MCF-6-f is the fast version of MCF-6, where the highly overlapped detection windows are eliminated after the first stage. There are the following observations: 1) MCF-6 and MCF-6-f both have the lower miss rates. Specifically, MCF-6 and MCF-6-f based on AlexNet have lower miss rates than MCF-2 by 2.79\% and 2.02\%, respectively. MCF-6 and MCF-6-f based on VGG16 have lower miss rates than MCF-2 by 4.21\% and 3.63\%, respectively. 2) MCF-6 and MCF-6-f is faster than MCF-2. For example, detection time of MCF-2 based VGG16 is 7.69s and that of MCF-6-f based on VGG16 is 1.89s. It means that detection speed of MCF-6-f is 4.07 times faster than that of MCF-2. 3) With little performance loss, MCF-6-f have faster detection speed than MCF-6. The loss of MCF-6-f based on AlexNet is 0.77\%, and the loss of MCF-6-f based on VGG16 is 0.58\%.

\subsection{Comparison with the state-of-the-Art}
In this section, MCF is based on HOG+LUV and all the five convolutional layers (i.e., C1-C5) in VGG16. Thus, MCF contains six-layer image channels. The features extracted in the first layer (i.e., HOG+LUV) are NNNF \cite{Cao_NNNF_arXiv_2015}, which is one of the state-of-the-art features. The features extracted in the remaining five layers (i.e., C1-C5) are zero-order feature (single pixel). Caltech10x is used for training the final classifier. To speedup the training process, negative samples are accumulated by five rounds of original NNNF, where the number of the trees in each round is 32, 128, 512, 2048, and 4096, respectively. The resulting classifier contains 4096 level-4 decision trees. S1 contains 2048 decision trees. S2-S6 each have 409 decision trees. In \cite{Zhang_RotatedFilters_arXiv_2016}, Zhang et al. provided a new, high quality ground truth for the training and test sets. The new annotations of Caltech10x is also used for training MCF. Original Caltech test set and new Caltech test set are both used for the evaluations.

\begin{figure}[!t]
\label{CaltechStandard}
\centering
\includegraphics[width=3in]{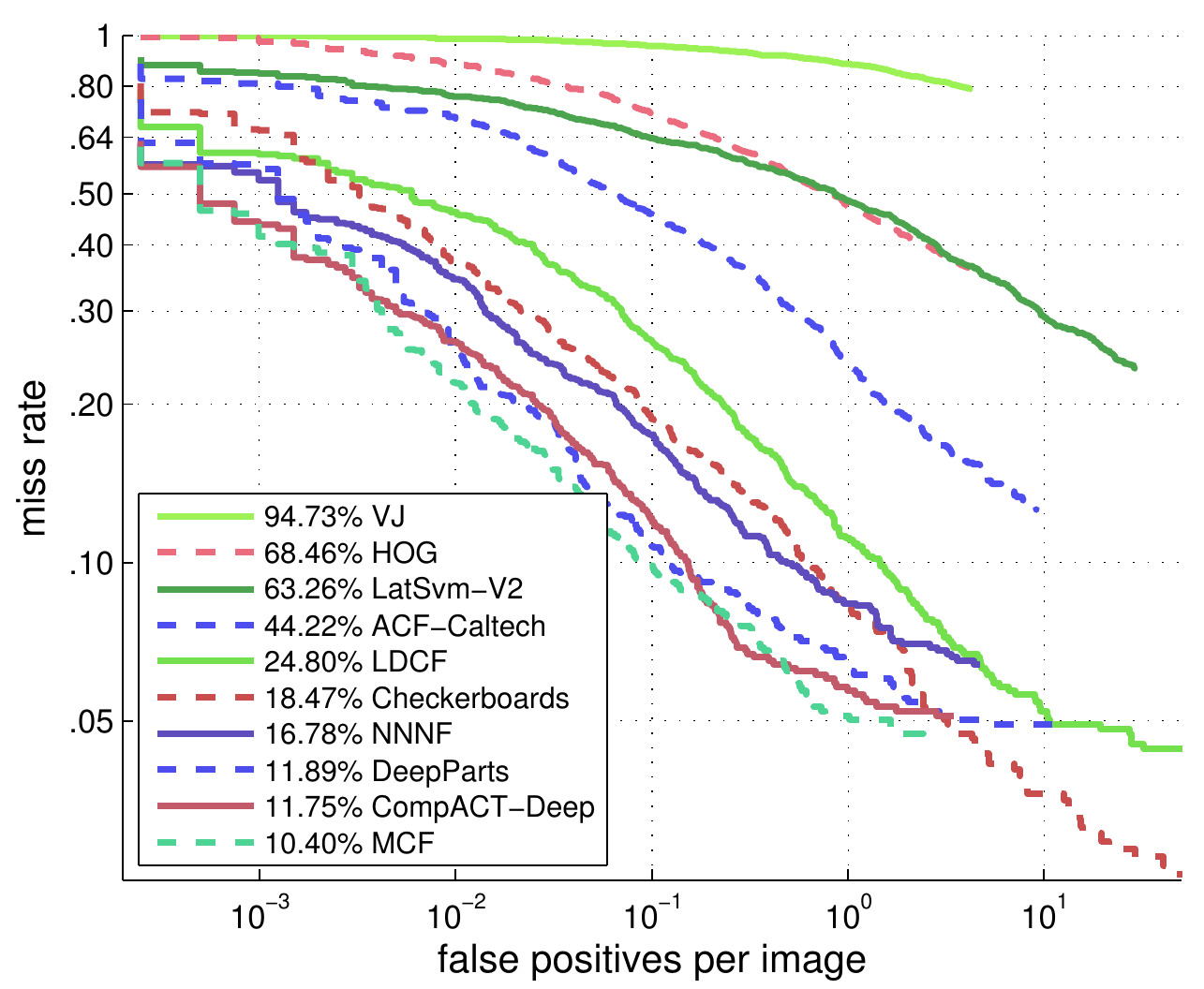}
\caption{ROC of Caltech test set (reasonable).} 
\end{figure}

Fig. 7 compares MCF with some state-of-the-art methods on the original annotations of the test set. ACF \cite{Dollar_ACF_PAMI_2014}, LDCF \cite{Nam_LDCF_NIPS_2014}, Checkboards \cite{Zhang_FCF_CVPR_2015}, NNNF \cite{Cao_NNNF_arXiv_2015}, DeepParts \cite{Tian_DeepParts_ICCV_2015}, and CompACT-Deep \cite{Cai_CompACT_ICCV_2015} are used. ACF \cite{Dollar_ICF_BMVC_2009} are trained on INRIA dataset \cite{Dalal_HOG_CVPR_2005}. The other methods are trained based on Caltech10x dataset. MCF achieves the state-of-the-art performance, which outperforms CompACT-Deep \cite{Cai_CompACT_ICCV_2015}, DeepParts \cite{Tian_DeepParts_ICCV_2015}, NNNF \cite{Cao_NNNF_arXiv_2015}, and Checkboards \cite{Zhang_FCF_CVPR_2015} by 1.35\%, 1.49\%, 6.38\%, and 8.07\%, respectively.

\begin{figure}[!t]
\label{RunTime}
\centering
\includegraphics[width=3.2in]{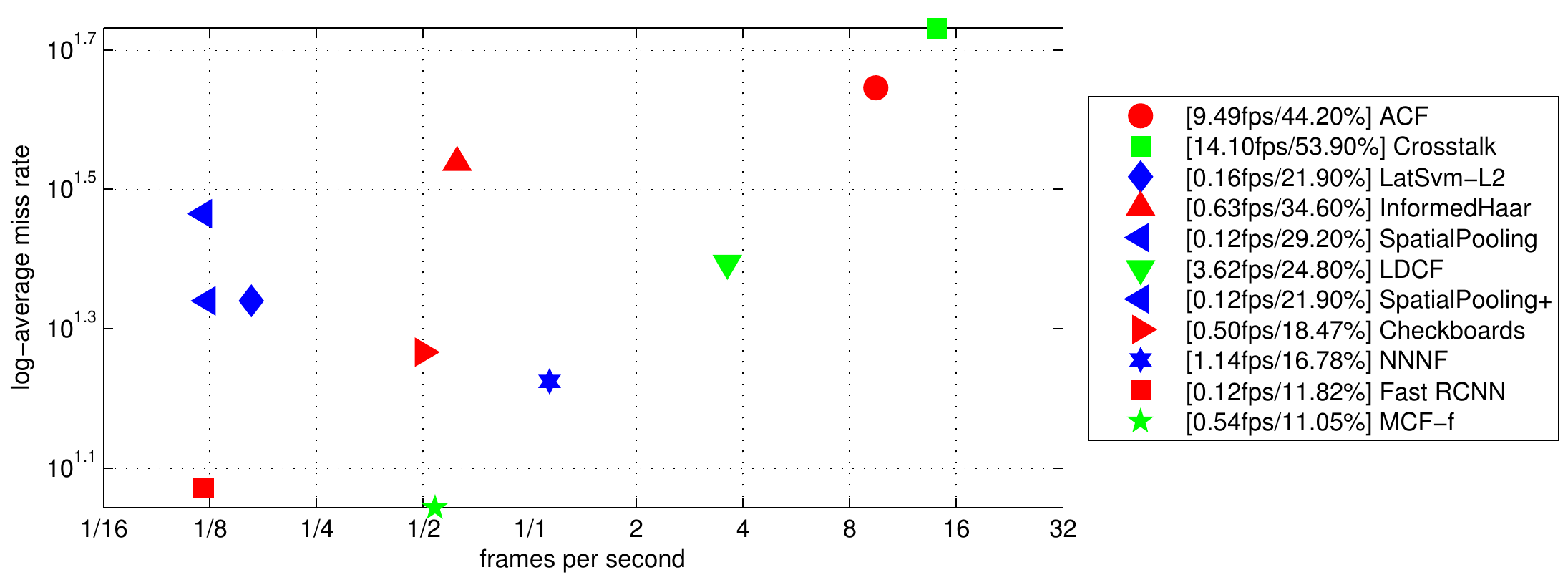}
\caption{Miss rates and FPS on Caltech pedestrian dataset are shown. Detection time of the methods are all tested on the common CPU (i.e., Intel Core i7-3700).} 
\end{figure}

Miss rates and Frames Per Second (FPS) of some methods based on CNN are visualized in Fig. 8. Detection time of the methods are all tested on the common CPU (i.e., Intel Core i7-3700). The best choice is that the miss rate is as small as possible while FPS is as large as possible. Though ACF \cite{Dollar_ACF_PAMI_2014} has very fast detection speed (9.49 fps), miss rate of ACF are very large. Fast RCNN reported in \cite{Li_ScaleAware_arXiv_2015} has the better performance (11.82\%), but the speed is very slow. MCF-f is the fast version of MCF with little performance loss (0.65\%). Compared to Fast RCNN \cite{Li_ScaleAware_arXiv_2015}, MCF is 4.5 times faster than it, while MCF has also lower miss rate than it by 0.77\%. Therefore, MCF has a better trade-off between detection time and detection performance.

\begin{figure}[!t]
\label{CaltechNew}
\centering
\includegraphics[width=3in]{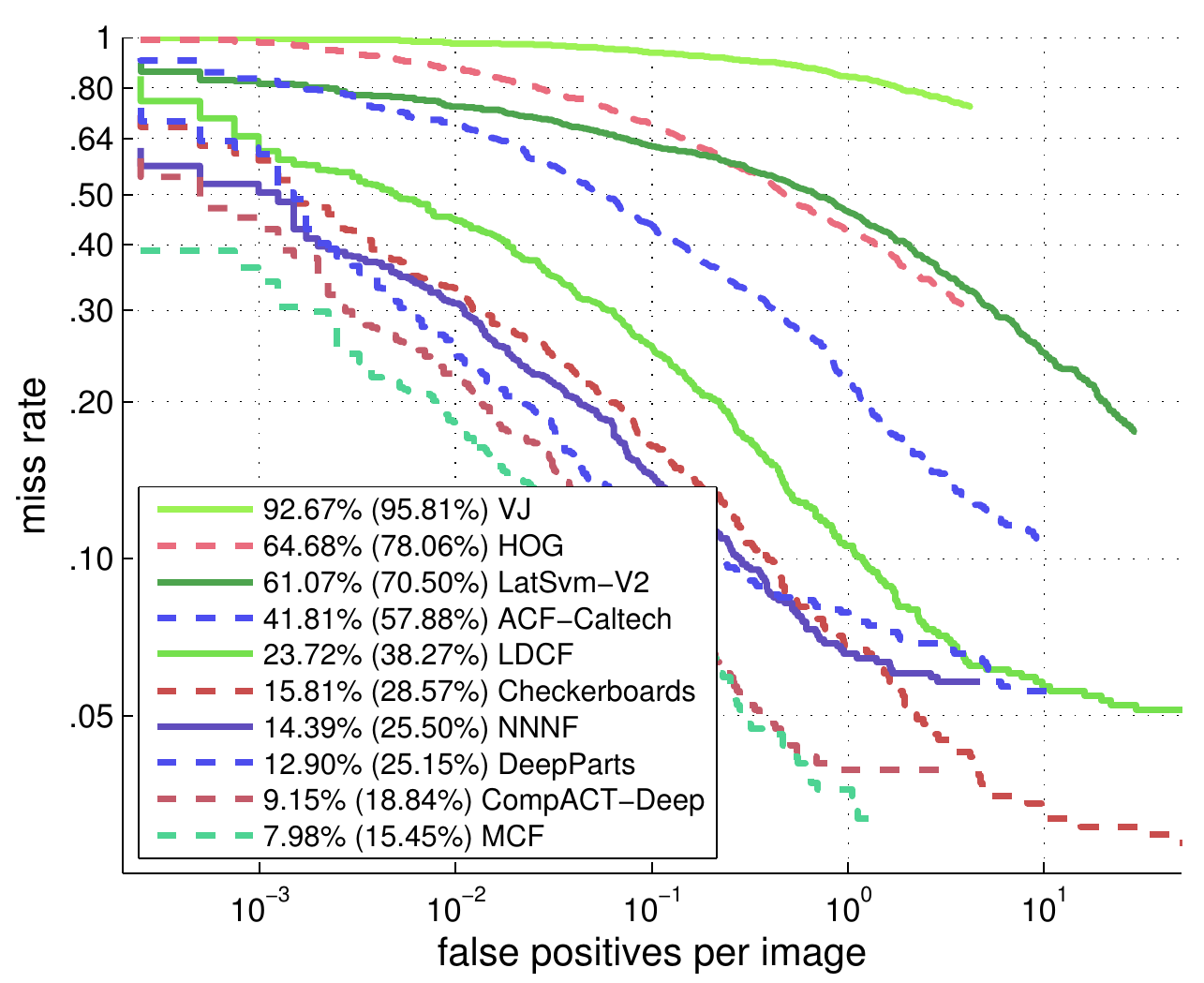}
\caption{ROC of Caltech test set using the new and accurate annotations \cite{Zhang_RotatedFilters_arXiv_2016}. Miss rates log-averaged over the FPPI range of [$10^{-2}$,$10^0$] and the FPPI range of [$10^{-4}$,$10^0$] are shown. They are represented by $MR_{-2}$ and $MR_{-4}$, respectively. $MR_{-2}$ ($MR_{-4}$) are shown in the legend.} 
\end{figure}

Based on the new and accurate annotations of the Caltech test set, Fig. 9 compares MCF with some state-of-the-art methods: CompACT-Deep \cite{Cai_CompACT_ICCV_2015}, DeepParts \cite{Tian_DeepParts_ICCV_2015}, Checkboards \cite{Zhang_FCF_CVPR_2015}, and NNNF \cite{Cao_NNNF_arXiv_2015}. Miss rates log-averaged over the FPPI range of [$10^{-2}$,$10^0$] and the FPPI range of [$10^{-4}$,$10^0$] are shown. They are represented by $MR_{-2}$ and $MR_{-4}$. $MR_{-2}$ ($MR_{-4}$) are shown in the legend. MCF is trained based on the Caltech10x with the new annotations. $MR_{-2}$ and $MR_{-4}$ of MCF achieve 7.98\% and 15.45\%, respectively. They are superior to all the other methods. Specifically, $MR_{-2}$ of MCF is 1.17\%, 4.92\%, and 6.41\% lower than that of CompACT-Deep \cite{Cai_CompACT_ICCV_2015}, DeepParts \cite{Tian_DeepParts_ICCV_2015}, and NNNF \cite{Cao_NNNF_arXiv_2015}. Compared to $MR_{-2}$ of MCF, $MR_-4$ of MCF has the better performance. Specifically, $MR_{-4}$ of MCF is 3.39\%, 9.70\%, and 10.05\% lower than that of CompACT-Deep \cite{Cai_CompACT_ICCV_2015}, DeepParts \cite{Tian_DeepParts_ICCV_2015}, and NNNF \cite{Cao_NNNF_arXiv_2015}. It means that MCF stably outperforms the other state-of-the-art methods.

\section{Conclusion}
In this paper, we have proposed a unifying framework, which is called Multi-layer Channels Features (MCF). Firstly, the handcrafted image channels and the layers in CNN construct the multi-layer image channels. Then a multi-stage cascade are learned from the features extracted in the layers, respectively. The weak classifiers in each stage are learned from the corresponding layer. On the one hand, due to the much more abundant candidate features, MCF achieves the state-of-the-art performance on Caltech pedestrian dataset (i.e., 10.40\% miss rate). Using the new and accurate annotations of the Caltech pedestrian dataset, miss rate of MCF is 7.98\%, which is superior to other methods. On the other hand, due to the cascade structure, MCF rejects many detection windows by the first few stages and then accelerates the detection speed. To further speedup the detection, the highly overlapped detection windows are eliminated after the first stage. Finally, MCF with VGG16 can run on the CPU by 0.54 fps.

\ifCLASSOPTIONcaptionsoff
  \newpage
\fi

\end{document}